\documentclass[letterpaper, 10 pt, journal, twoside]{IEEEtran}

\usepackage[pagebackref=false,breaklinks=true,colorlinks,bookmarks=false]{hyperref}
\usepackage{url}
\usepackage[numbers, sort]{natbib} %

\usepackage{graphics}           
\usepackage{times}              
\usepackage{amsmath}            
\usepackage{amssymb}            
\usepackage{graphicx}
\usepackage{algorithm}
\usepackage[noend]{algpseudocode}
\usepackage{booktabs}
\usepackage{color}
\definecolor{instructioncolor}{rgb}{.5,.5,.5}

\usepackage[font=small]{caption}

\def\secref#1{Sec.~\ref{#1}}
\def\figref#1{Fig.~\ref{#1}}
\def\tabref#1{Tab.~\ref{#1}}
\def\eqref#1{Eq.~(\ref{#1})}

\makeatletter
\usepackage{xspace}
\DeclareRobustCommand\onedot{\futurelet\@let@token\@onedot}
\def\@onedot{\ifx\@let@token.\else.\null\fi\xspace}
\def\eg{e.g\onedot} 
\def\ie{i.e\onedot}

\def\etal{{et al}\onedot}
\makeatother

\def\etalcite#1{\etal~\cite{#1}}

\usepackage{array}
\newcolumntype{L}[1]{>{\raggedright\let\newline\\\arraybackslash\hspace{0pt}}m{#1}}
\newcolumntype{C}[1]{>{\centering\let\newline\\\arraybackslash\hspace{0pt}}m{#1}}
\newcolumntype{R}[1]{>{\raggedleft\let\newline\\\arraybackslash\hspace{0pt}}m{#1}}

\newcommand{\RR}{\mathbb{R}}

\newcommand{\abs}[1]{|#1|}

\renewcommand{\b}[1]{\mbox{\boldmath$#1$}}

\renewcommand{\v}[1]{{\b #1}} 

\newcommand{\m}[1]{{\mbox{{\sffamily\slshape{#1\/}}}}}

\usepackage{subfigure}
\usepackage{pifont}
\usepackage{amssymb}
\usepackage{multirow}
\usepackage{graphicx}
\usepackage[table]{xcolor}

\newcommand{\cmark}{\ding{51}}%

\renewcommand{\thepage}{} %
\definecolor{lightgray}{rgb}{0.83, 0.83, 0.83}

\title{\LARGE \bf Efficient Spatial-Temporal Information Fusion \\ for LiDAR-Based 3D Moving Object Segmentation}
{}

\author{Jiadai Sun \quad Yuchao Dai$^{*}$ \quad Xianjing Zhang \quad Jintao Xu \quad Rui Ai \quad Weihao Gu \quad Xieyuanli Chen
  \thanks{J. Sun and Y. Dai are with Northwestern Polytechnical University, China.}
  \thanks{X. Zhang, J. Xu, R. Ai and W. Gu are with HAOMO.AI Tech. Co., Ltd.}
  \thanks{$^*$ corresponding author: daiyuchao@gmail.com}
  \thanks{This work has partially been funded by the National Key Research and Development Program of China under Grant 2018AAA0102803, and by the HAOMO.AI Technology Co. Ltd.}
}

\begin{document}
\maketitle
\thispagestyle{empty}
\pagestyle{empty}

\begin{abstract}
    Accurate moving object segmentation is an essential task for autonomous driving. It can provide effective information for many downstream tasks, such as collision avoidance, path planning, and static map construction.
    How to effectively exploit the spatial-temporal information is a critical question for 3D LiDAR moving object segmentation (LiDAR-MOS).
    In this work, we propose a novel deep neural network exploiting both spatial-temporal information and different representation modalities of LiDAR scans to improve LiDAR-MOS performance.
    Specifically, we first use a range image-based dual-branch structure to separately deal with spatial and temporal information that can be obtained from sequential LiDAR scans, and later combine them using motion-guided attention modules. We also use a point refinement module via 3D sparse convolution to fuse the information from both LiDAR range image and point cloud representations and reduce the artifacts on the borders of the objects.
    We verify the effectiveness of our proposed approach on the LiDAR-MOS benchmark of SemanticKITTI. Our method outperforms the state-of-the-art methods significantly in terms of LiDAR-MOS IoU.
    Benefiting from the devised coarse-to-fine architecture, our method operates online at sensor frame rate.   
    The implementation of our method is available as open source at: \url{https://github.com/haomo-ai/MotionSeg3D}.
\end{abstract}

\section{Introduction}
\label{sec:intro}

Environmental perception can help vehicles observe and understand the surrounding situation. The ability to recognize and distinguish between dynamic and static objects in the environment is the key to safe and reliable autonomous navigation. At the same time, this information can also be used for many downstream tasks, such as avoiding obstacles~\cite{peters2021rss}, static map construction~\cite{chen2021ral}, and path planning~\cite{kummerle2014jfr}.
Therefore, being able to perform accurate and reliable moving object segmentation (MOS) based on LiDAR sequences online is a key capability to improve the perception of autonomous mobile systems.
To reason about the motion of the surrounding objects, one needs to exploit 4D spatio-temporal information.

\begin{figure}[!t]
	\centering
	\includegraphics[width=\linewidth]{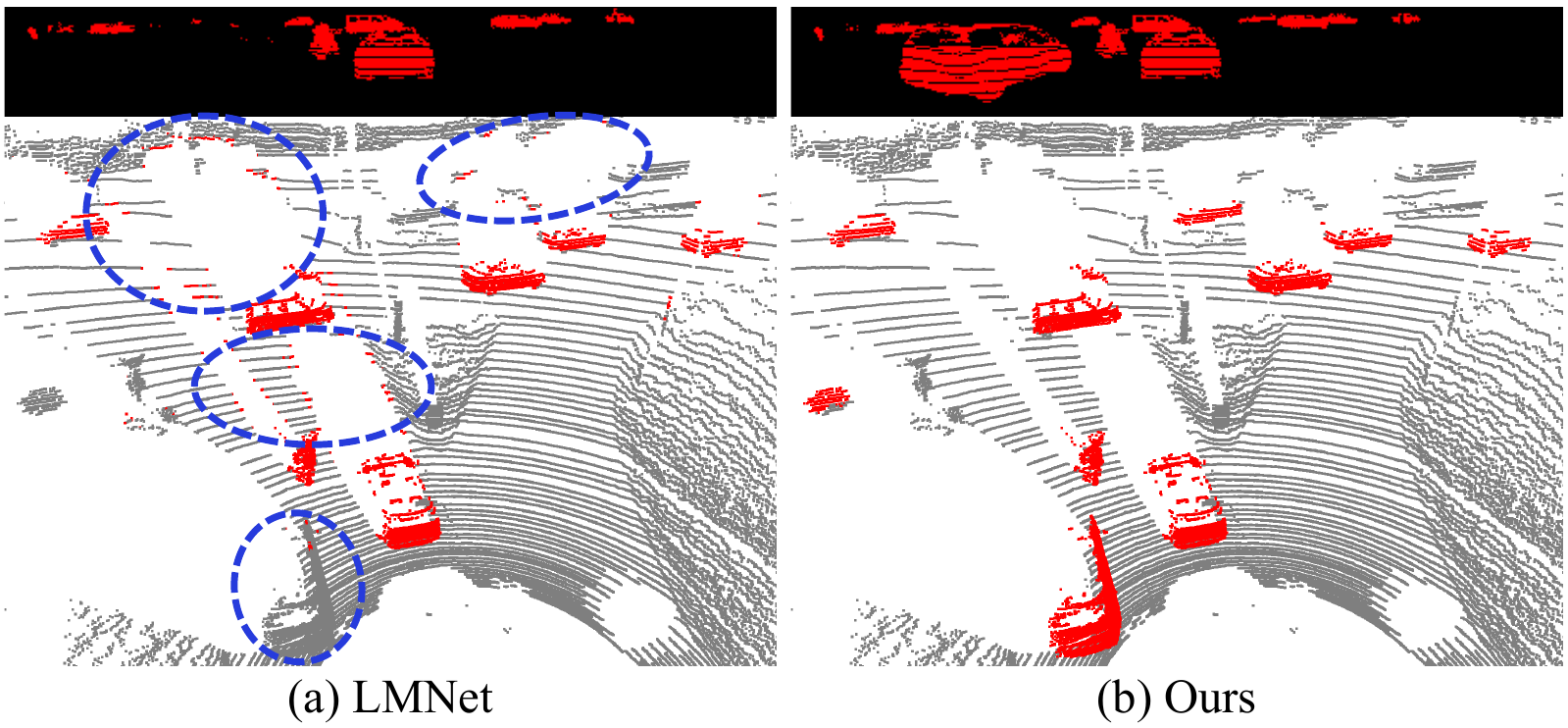}
	\caption{
    LiDAR-MOS comparison between our method and LMNet.
    The upper row shows the segmentation results on range images and the lower row shows the results in 3D point clouds. Red pixels/points are moving objects while black ones represent static objects. Blue circles highlight the wrong predictions.
	(a)~There are many artifacts on the borders of objects, but not obvious in the range image.
	(b)~Fusing spatial-temporal and different representation of LiDAR scans, our method reduces the artifacts and achieves \textit{sota} performance.}
	\label{fig:motivation}
 	\vspace{-0.2cm}
\end{figure}

MOS can be viewed as a higher-level two-class ``semantic" segmentation task. Instead of distinguishing the basic semantic classes, \eg, humans, vehicles, and buildings, MOS infers the dynamic properties of objects and separates the \emph{actually moving} objects, \eg, driving cars and pedestrians, from static or non-moving objects \eg, buildings and parked cars.
For point cloud segmentation, popular existing solutions can be divided into point cloud-based \cite{thomas2019iccv,hu2020randla}, voxel-based \cite{tang2020sparseconv, zhu2021cylinder3D}, and range image-based~\cite{milioto2019iros,li2022ral,cortinhal2020iv}.
Point-based methods can extract effective features from unordered point clouds, but they are difficult to scale effectively to large-scale point cloud data.
Sparse voxels convolution~\cite{choy20194dMinkowski, tang2020sparseconv} can reduce the computational burden of point clouds, but voxelization will introduce information loss. 
Range images are used as a comparably lightweight intermediate representation and attractive for online applications. However, it causes boundary-blurring issues due to back-projection.
Instead of using a single representation, we propose to first use a range image-based backbone to obtain a \emph{coarse} segmentation and then use a lightweight 3D voxel sparse convolution module to \emph{refine} the segmentation results. Using such a coarse-to-fine architecture, our method combines the advantages of different representation modalities of LiDAR scans to alleviate the boundary-blurring issue, while maintaining good efficiency. Example results are shown in~\figref{fig:motivation}.

Different from existing segmentation methods, which are done on a single LiDAR scan, determining whether an object is moving or not usually requires multi-frame observations. 
Chen~\etalcite{chen2021ral} propose LMNet to directly use off-the-shelf segmentation networks \cite{milioto2019iros,cortinhal2020iv,li2022ral}. 
It exploits the spatial-temporal information by simply concatenating the residual images calculated from multiple continuous scans.
In contrast to LMNet, we propose a novel dual-branch structure that first deals with spatial and temporal information separately and then fuses them using motion-guided attention modules.

The main contribution of this paper is a novel deep neural network to tackle online LiDAR-MOS in 3D data. Our method uses a dual-branch structure bridged by motion-guided attention modules to exploit spatial-temporal information from sequential LiDAR scans.
We use a coarse-to-fine architecture fusing range-image and point-cloud representations to reduce the artifacts on the borders of the objects without applying a kNN post-processing and a semantic refinement. Based on that, our method achieves online performance, \ie, performs faster than the frame rate (10\,Hz) of a typical 3D LiDAR sensor.
Our method achieves the state-of-the-art LiDAR-MOS performance on the SemanticKITTI-MOS benchmark~\cite{chen2021ral}.
When using the proposed extra data from KITTI-road sequences, our method gains around~10\% improvement on the hidden test of the benchmark. We will release the extra annotated data of the KITTI-road dataset together with the implementation of our method to support future research. 

Our contributions can be summarized as follows:
(i) We propose a dual-branch structure bridged by motion-guided attention modules to better exploit the temporal motion information in residual images.
(ii) We use a coarse-to-fine architecture to reduce blurred artifacts on object borders.
(iii) Our method achieves the state-of-the-art performance in LiDAR-MOS on the SemanticKITTI-MOS benchmark.

\section{Related Work}
\label{sec:related}

\begin{figure*}[!t]
	\centering
	\includegraphics[width=\linewidth]{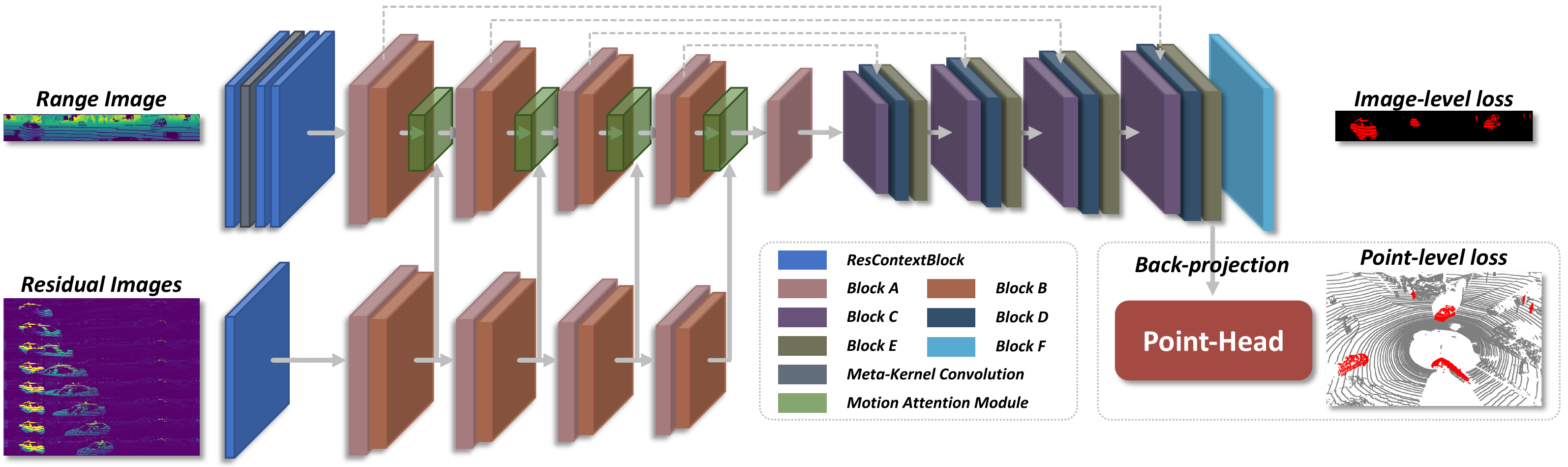}
	\caption{Overview of our method. We extend and modify SalsaNext~\cite{cortinhal2020iv} into a dual-branch and dual-head architecture, consisting of a range image branch (Enc-A) to encode the appearance feature, a residual image branch (Enc-M) to encode the temporal motion information, and use multi-scales motion guided attention module to fuse them. And then an image head with skip connections is used to decode the features from fronts. Finally, we back-project 2D features to 3D points and use a point head to further refine the segmentation results. 
	Specifically, BlockA and BlockE are the ResBlocks with dilated convolution, BlockB is the pooling and optional dropout layer, BlockC is the PixelShuffle and optional dropout layer, BlockD is the skip connection with optional dropout, BlockF is the fully connected layer.
	} \label{fig:overview}
\end{figure*}

Moving object segmentation~(MOS) has been well studied in the literature using image sequences~\cite{yao2020tist} or \mbox{RGB-D} data~\cite{maddalena2018joi}. However, it is still challenging for LiDAR data due to the sparsity and uneven distribution of the range measurements.
Also, how to exploiting 4D spatio-temporal information from point cloud sequence is still an open question.
Here, we focus on approaches using only LiDAR sensors and refer for visual approach to existing surveys~\cite{maddalena2018joi, yao2020tist}. 

There are some geometric-based methods~\cite{yoon2019cvr, dewan2016iros} to tackle the LiDAR-MOS problem, they do not need training data and procedure but occasionally result in incomplete or inaccurate detection of moving objects.
There are map cleaning-based methods that can be used to separate moving objects from static LiDAR maps. For example, Kim~\etalcite{kim2020iros} %
exploit the consistency check between the query scan and the pre-built map to remove dynamic points.
The map is then refined using a multi-resolution false prediction reverting algorithm.
Lim~\etalcite{lim2021ral} remove dynamic objects by checking the occupancy of each sector of LiDAR scans and then revert the ground plane by region growth. In contrast, Arora~\etalcite{arora2021ecmr} segment the ground plane and then remove the ``ghost effect'' caused by the moving object during mapping.
Chen~\etalcite{chen2022arxiv} use the map cleaning method with clustering and multi-object tracking to track the trajectories of different objects and generate training labels for LiDAR-MOS based on the tracking results. 
Even though such map cleaning methods can distinguish moving and static objects, they usually can only run offline and are not suitable for online MOS.

For online LiDAR-MOS, there are deep network-based methods, which use generic end-to-end trainable models to learn local and global statistical relationships directly from data.
For example, there are point cloud scene flow methods~\cite{baur2021iccv, liu2019cvpr,wu2020pointpwc}, which usually estimate motion vectors between two consecutive scans. Based on the predicted motion vectors, it separates moving and non-moving objects by estimating the velocity of every point, which may not differentiate between slowly moving objects and sensor noise. 
It is worth mentioning that most of them are hard to handle large scans (about 100k points) obtained by the LiDAR sensor, and the real-time performance is difficult to guarantee.
In addition, it is also possible to determine whether the object is moving according to the displacement of the bounding box, which requires some prior information for target detection or tracking~\cite{beltran2018birdnet, zhou2018voxelnet, lang2019pointpillars, vaquero2020dual, shi2020pv}.

Semantic segmentation can be viewed as a related step towards MOS.
Recently, LiDAR-based semantic segmentation methods operating only on sensor data have achieved great success~\cite{milioto2019iros, cortinhal2020iv,li2022ral,thomas2019iccv,tang2020eccv,shi2020cvpr}.
However, most single LiDAR-frame semantic segmentation methods only find \textit{movable} objects, \eg, vehicles and humans, but do not distinguish between \emph{actually moving} objects, such as walking pedestrians or driving cars, and non-moving/static objects, like building structures or parked cars.
Wang~\etalcite{wang2012icra} also tackle the problem of segmenting things that could move from 3D laser scans of urban scenes, \eg, cars, pedestrians, and bicyclists.
Ruchti~\etalcite{ruchti2018icra} use a learning-based method to predict the probabilities of potentially movable objects.
Based on the semantic segmentation results~\cite{milioto2019iros}, Chen~\etalcite{chen2019iros} propose a semantic LiDAR SLAM, which detects and filters out moving objects by checking the semantic consistency between online observation and semantic map representation.

In contrast to the single-frame methods, some methods~\cite{thomas2019iccv, tang2020eccv, shi2020cvpr} operate on multiple point clouds or an aggregated point cloud submap to achieve better segmentation results and at the same time separate moving and non-moving objects. 
However, these methods perform operations directly on point clouds, which are often laborious and difficult to train.
Furthermore, most of them are both time-consuming and resource-intensive, which might not be applicable for autonomous driving.

The most related work to ours is LMNet~\cite{chen2021ral}, which also separates moving and non-moving objects using LiDAR scans. 
Instead of designing a new network structure, it reuses the off-the-shelf LiDAR semantic segmentation methods~\cite{milioto2019iros,li2022ral,cortinhal2020iv}.
To obtain inter-frame motion information, it feeds the multi-frame residual images directly into the existing structure as extra channels to the range image.
Such simple concatenation without special design often can not accurately exploit the motion information contained in the spatio-temporal scan sequence.
Moreover, LMNet only uses range image representation, bringing many artifacts during back-projection to point clouds, as shown in~\figref{fig:motivation}. Different from LMNet, we propose a novel network. It uses two specific branches to extract appearance features from range images and temporal motion features from residual images, respectively. 
And then, it uses motion-guided attention modules with different scales to fuse them. 
In the final stage of decoding, we back-project the 2D features to the 3D point cloud, and use a lightweight 
sparse convolution module to refine the segmentation results.
\section{Our Approach}
\label{sec:methods}

\subsection{Preliminaries}
\noindent \textbf{Range Image Representation.}\label{sec:range} It is a lightweight data representation obtained by projecting the 3D point cloud into 2D space.
The advantages of range representation are that it can alleviate the massive consumption caused by the direct processing of point cloud data, and it can facilitate the use of mature 2D convolutional neural networks that have been well studied in vision-based tasks.
The range image is widely adopted in various tasks \cite{li2022ral, chen2021icra, cortinhal2020iv, milioto2019iros, fan2021rangedet}, and we only give a quick review.
For each LiDAR point $\v{p}=(x, y, z)$ with Cartesian coordinates, a spherical mapping $\Pi: \RR^3 \mapsto \RR^2$ is used to transform it to image coordinates, as following,
\begin{align}
	\left( \begin{array}{c} u \vspace{0.0em} \\ v \end{array}\right) & = \left(\begin{array}{cc} \frac{1}{2}\left[1-\arctan(y, x) \, \pi^{-1}\right]~\,~w   \vspace{0.5em} \\
			\left[1 - \left(\arcsin(z\, r^{-1}) + \mathrm{f}_{\mathrm{up}}\right) \mathrm{f}^{-1}\right] \, h\end{array} \right), \label{eq:projection}
\end{align}
where $(u,v)$ are image coordinates, $(h, w)$ are the height and width of the desired range image, $\mathrm{f}~{=}~\mathrm{f}_{\mathrm{up}}~{+}~\mathrm{f}_{\mathrm{down}}$ is the vertical field-of-view of the sensor, and $r~{=}~||\v{p}||_2$ is the range of each point.
After that, we can use $(u,v)$ to index the 3D point and integrate its coordinates $(x, y, z)$, range~$r$, and intensity~$e$ as the five channels of the range image.

\vspace{1mm}
\noindent \textbf{Residual Images.}\label{sec:residual}
We follow LMNet~\cite{chen2021ral} using residual images to exploit the spatial-temporal information from sequential LiDAR scans. 
To generate a residual image between the current frame $l$ and the previous frame $k$, there are three steps. First, using the relative pose to transform the previous scans $k$ to the current coordinate system. Second, re-projecting the transformed past LiDAR points into range image. Third, computing the residual $d^{l}_{k,i}$ for each pixel $i$ by computing the normalized absolute difference between the ranges of the current frame $l$ and the transformed frame $k$ as
\begin{align}
	d^{l}_{k,i} &= 
        \begin{cases}
		{\abs{r_i - r^{k \rightarrow l}_i}}/{r_i} & \text{$i \in$ valid pixels,} \\
        0 & \text{otherwise,}
        \end{cases}
\end{align}
where $r_i$ is the range value of $\v{p}_{i}$ from the current frame located at image coordinates $(u_i,v_i)$ and $r^{k \rightarrow l}_i$ is the corresponding range value from the transformed scan located at the same image pixel. Please refer to \cite{chen2021ral} for more details.

\begin{figure}[t]
	\centering
	\includegraphics[width=\linewidth]{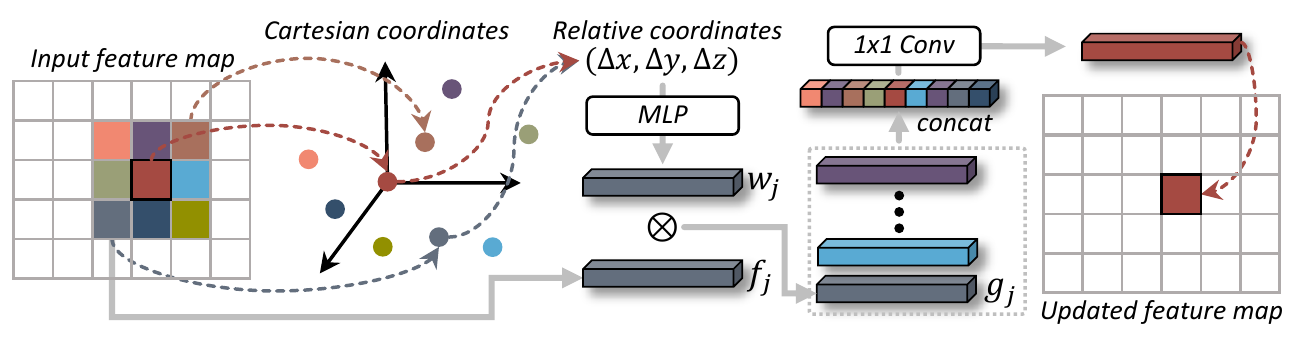}
	\caption{Architecture of the Meta-Kernel Module~\cite{fan2021rangedet}. According to the 3D coordinates stored in the range image and the input feature map, the weight of $3\!\times\!3$ neighborhood can be calculated via the relative coordinates of the center point, and then a $1\!\times\!1$Conv is used to aggregate neighbor features to update the center feature.}
	\label{fig:net_metakernel}
\end{figure}

\vspace{1mm}
\noindent \textbf{Meta-Kernel Convolution.}\label{sec:metakernel}
As argued by Fan~\etalcite{fan2021rangedet}, using 2D convolution on range images cannot fully exploit the 3D geometric information due to the dimensionality reduction of the spherical projection.
Exploiting meta-kernel convolution, we can take advantage of the 3D geometric information by using the relative Cartesian coordinates of the $3\!\times\!3$ neighbors of the center, as shown in \figref{fig:net_metakernel}.
A shared MLP takes these relative coordinates as input to generate nine weight vectors $w_j$, and does element-wise product on the corresponding nine feature vectors $f_j$.
Finally, by passing a concatenation of the nine neighbors output $g_j$ to a $1\!\times\!1$ convolution, we aggregate the information from different channels and different sampling locations to update the center feature vector.

\subsection{Network Overview}
We assume a given sequence of LiDAR scans $\{\m{S}_{t}\}_{t=1}^{T}$ and poses $\left\{\xi_{t}\right\} \in \mathbb{SE}(3)$ provided by a SLAM system, where $t$ represents the time step. 
The goal is to get accurate point-wise segmentation of moving objects for the current frame, using only the current and previous LiDAR scans.
The system architecture is illustrated in \figref{fig:overview}. We propose to use both 2D range images and 3D point clouds to obtain accurate 3D segmentation. Our method is mainly based on range images and refined by a lightweight 3D point cloud network. 

Our proposed network architecture is built upon the SalsaNext~\cite{cortinhal2020iv}, a single encoder-and-decoder network for LiDAR range image-based semantic segmentation. 
To make it suitable for MOS, we extend and modify it into a dual-branch and dual-head network, consisting of a range image branch (Enc-A) to encode the appearance feature, a residual image branch (Enc-M) to encode the temporal motion information, an image head (ImageHead) with skip connections to decode the features from both Enc-A and Enc-M, and a point head (PointHead) to further refine the segmentation results.
Specifically, in the feature encoding stage, we first use the meta-kernel operator to better capture the 3D spatial information, and then use a motion-guided attention module to more effectively fuse the motion information extracted from residual images. In the final stage of decoding, besides the loss on the range image dealt with the ImageHead, we also back-project the 2D features to 3D point clouds, and use a lightweight sparse convolution module (PointHead) to refine the segmentation results.

\subsection{Dual Branches with Motion Guided Attention Module}
Different from LMNet~\cite{chen2021ral}, which directly concatenates the range image and residual images together as the input of the original SalsaNext, we use two specific branches to extract appearance features from the range images and motion features from the residual images, respectively. 
To preserve descriptive features, we furthermore replace the average pooling in BlockB with SoftPool as suggested by~\cite{stergiou2021softpool}.
In Enc-A, we place one Meta-Kernel convolution layer after the first ResContextBlock~\cite{cortinhal2020iv} to learn dynamic weights from relative Cartesian coordinates and enable the network to obtain more geometric information from the range images, making the convolution more suitable to the range images.
Different from the range image branch, in \mbox{Enc-M}, we only use one ResContextBlock to avoid overfitting of the residual images.

\begin{figure}[t]
	\centering
	\includegraphics[width=\linewidth]{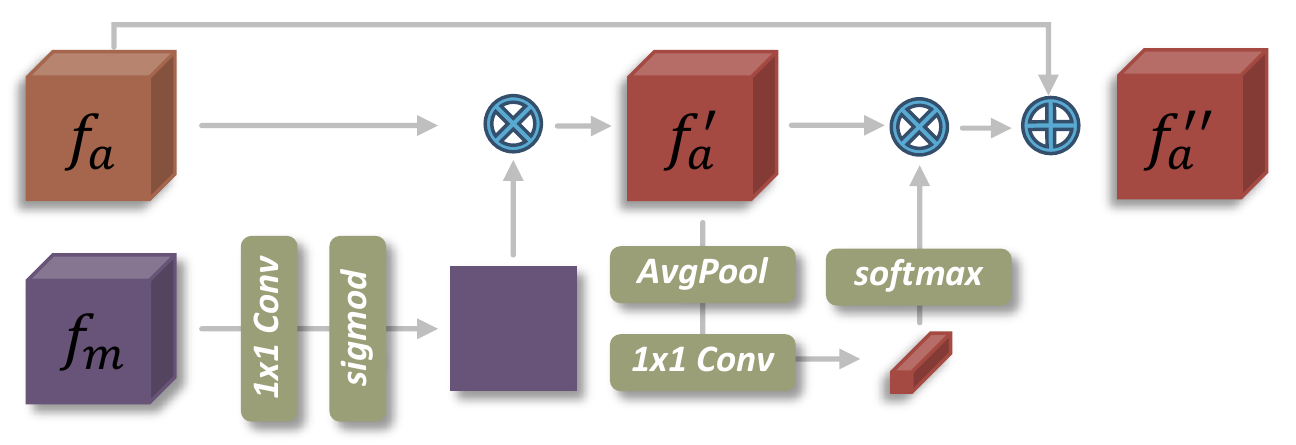}
	\caption{Architecture of the Motion Attention module. The spatial attention and channel attention are used to fuse the moving feature from residual image and appearance feature from range image.
	}
	\vspace{-0.2\baselineskip}
	\label{fig:net_attention}
\end{figure}

Inspired by video-based object segmentation methods \cite{li2019motion_attention,sun2021munet,dave2019towards} using optical flow to obtain appearance features, we add a spatial and channel attention module~\cite{li2019motion_attention} to extract motion information from residual images. Such motion information enhances the appearance features extracted from range images, \ie, exploiting motion information and emphasizing more salient areas in appearance features.
As illustrated in \figref{fig:net_attention}, we use a similar structure to Li~\etalcite{li2019motion_attention} to fuse the features of the two branches. 
We denote $f_a$ as the appearance feature of range images from Enc-A and $f_m$ as the motion feature of residual images from Enc-M branch, and have:
\begin{align}
	f_{a}^{\prime} &\!=\!f_{a} \otimes \operatorname{Sigmoid}\left(\operatorname{Conv_{1\times1}}\left(f_{m}\right)\right), \\
	f_{a}^{\prime \prime} &\!=\!f_{a}^{\prime} \otimes\left[\operatorname{Softmax}\left(\operatorname{Conv_{1\times1}}\left(\operatorname{APool}\left(f_{a}^{\prime}\right)\right)\right) \cdot C\right]+f_{a},
\end{align}
where all $f$ represent feature map of size $C\times h \times w$. $\operatorname{APool}(\cdot)$ denotes average pooling in the spatial dimensions. Our method first uses a spatial attention to emphasize the spatial locations on the current appearance feature $f_a$ using the motion feature $f_m$ and generates a motion-salient feature $f_a^{\prime}$. Second, we adopt the channel-wise attention to strengthen the responses of essential attributes by channel-wise attention and generate the final spatial-temporal fused feature $f_a^{\prime \prime}$.

\begin{figure}[!t]
	\centering
	\includegraphics[width=\linewidth]{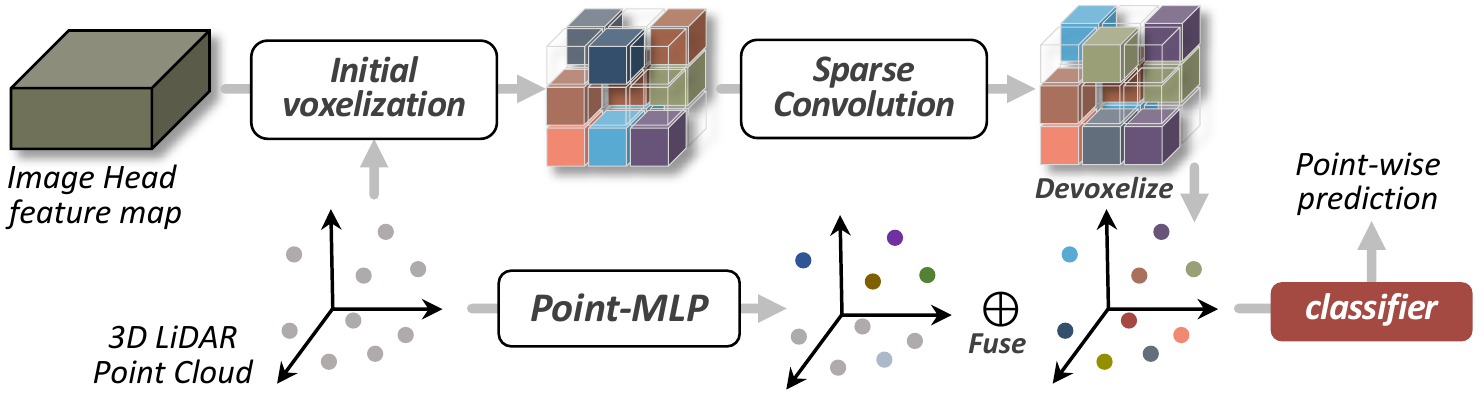}
	\caption{Architecture of the PointHead module. According to the index of spherical projection, 2D features are back-projected to 3D point, and then we use a sparse voxel-based branch and point-based branch to extract point-wise features for more accurate classification. The upper row represents the voxel-branch and the lower is point-branch.}
	\vspace{-0.2\baselineskip}
	\label{fig:net_pointhead} %
\end{figure}

\subsection{Coarse-to-Fine: Point Refine Module via 3D SparseConv}

Although LMNet~\cite{chen2021ral} can perform LiDAR-MOS only using a 2D segmentation network, the boundary-blurring effect is unavoidable due to the limited resolution of the feature maps and the dimensionality reduction of the range image representation. %
This leads to false positive predictions around object boundaries.
To tackle this issue, we propose a coarse-to-fine strategy.
Instead of only using the pixel-wise loss, we propose a point head to refine the segmentation results after the 2D convolutional network. This two-step coarse-to-fine strategy makes the supervision more effective and utilizes both pixel-wise and point-wise supervision.
To this end, we back-project/re-index the feature maps of the last layer to the original point locations. Then, we refine point-wise segmentation results by combining this with spatial information performing a relatively lightweight point cloud convolution operation. 
Back-projection is performed with indices computed when the point cloud is projected to the 2D range image. %
As shown in \figref{fig:net_pointhead}, we back-project 2D feature maps $(C \times h \times w)$ from ImageHead into point-wise features $(N \times C)$. Then, we use points in Cartesian coordinates $(x,y,z)$ and point-wise features to initially voxelize 3D sparse voxel. Finally, we use two branches of 3D sparse convolution and point cloud-based MLP to further refine the results using spatial geometry, and reduce the artifacts that occur around the object boundaries.

\subsection{Loss Functions}
Following \cite{cortinhal2020iv,chen2021ral}, two loss functions are used to supervise our network.
The total loss function combines both weighted cross-entropy and Lov\'{a}sz-Softmax losses~\cite{berman2018lovasz} as $\mathcal{L} = \mathcal{L}_{wce} + \mathcal{L}_{ls}$.
To alleviate the imbalanced distribution over different classes, the cross-entropy loss function is weighted with inverse square root frequency for each class, defined as
\begin{equation}
    \mathcal{L}_{wce}(y, \hat{y})=-\sum \alpha_{i} p\left(y_{i}\right) \log \left(p\left(\hat{y}_{i}\right)\right), \ \alpha_{i}=1 / \sqrt{f_{i}},
\end{equation}
where ${y}_{i}$ and $\hat{y}_{i}$ are the true and predicted labels and $f_i$ is the frequency of the $i^{th}$ class.
The Lov\'{a}sz-Softmax loss can be formulated as follows:
\begin{equation}
    \mathcal{L}_{ls}=\frac{1}{|C|} \sum_{c \in C} \overline{\Delta_{J_{c}}}(m(c)),\  m_{i}(c)\!=\!\left\{\begin{array}{ll}1\!-\!x_{i}(c) & \text{if } c\!=\!y_{i}(c) \\
x_{i}(c) & \text{otherwise}\end{array}, \right.
\end{equation}
where $|C|$ is the class number, $\overline{\Delta_{J_{c}}}$ represents the Lov\'{a}sz extension of the Jaccard index, $x_{i}(c) \in[0,1]$ and $y_{i}(c) \in \{-1,1\}$ hold the predicted probability and ground truth label of pixel~$i$ for class $c$, respectively.

The same loss function is applied to both pixel and point levels in a two-stage training scheme. We first apply the loss on the pixel level to supervise the range image-based encoder-and-decoder backbone. 
Then, we freeze the 2D range image network and apply the loss point-wisely to train the proposed PointHead. In this way, we train the network fusing supervision from both range image and point cloud representations.

\subsection{Implementation Details}
We use PyTorch \cite{paszke_pytorch_NIPS_2019} library to implement our method, which is trained with 4 NVIDIA RTX 3090 GPUs.
The size of the range image is set to $64 \times 2048$. 
We apply the same data augmentation as used in LMNet~\cite{chen2021ral} during training.
We minimize $\mathcal{L}_{wce}$ and $\mathcal{L}_{ls}$ using the stochastic gradient descent %
with momentum 0.9 and weight decay 0.0001. The initial learning rate is set to 0.01. %
We use the implementation of 3D sparse convolution from TorchSparse~\cite{tang2020sparseconv} to implement our PointHead.
We train the network using a two-stage training scheme. First, we train the 2D convolutional network with the image labels. After that, we freeze the parameters of the 2D encoder-decoder network and use the point cloud labels to train the PointHead separately.

\section{Experiments}\label{sec:exp} %
In this section, we conduct a series of experiments on the SemanticKITTI-MOS dataset~\cite{chen2021ral} to evaluate the quality of the MOS and different design considerations of our method.

\subsection{Experiment Setups}
\textbf{Datasets.} We train and evaluate our method on the SemanticKITTI-MOS dataset~\cite{chen2021ral}, which uses the same split for training and test as used in the original odometry dataset and remapping all the 28 semantic classes into only two types: moving and non-moving/static objects.
The dataset contains 22 sequences in total, where 10 sequences (19,130 frames) are split for training, 1 sequence (4,071 frames) for validation, and 11 sequences (20,351 frames) for testing. 

There are currently only a few datasets available for 3D LiDAR-based MOS, and the ratio of moving objects in the current Semantic-KITTI MOS dataset is relatively small. 
We call a LiDAR scan a dynamic frame if the number of moving points in that frame is larger than 100, otherwise it is a static frame. The proportion of dynamic frames is only $25.77\%$ in the train split.
To have more data at hand, we use an automatic label generation method~\cite{chen2022arxiv} to first automatically generate coarse labels for the KITTI-road dataset~\footnote{\url{http://www.cvlibs.net/datasets/kitti/raw_data.php?type=road}} and then manually refine them to enrich the training data for this task. We label 12 sequences of KITTI-road, where 6 sequences (2,905 frames) are used for training and 6 sequences (2,889 frames) for validation. We will release this additional labeled dataset to facilitate further research.

Because of the unequal distribution of dynamic and static training samples, as also indicated in~\cite{mohapatra2021limoseg}, we omit frames from continuous static frames to speed up the training, \ie, using a smaller downsampled dataset for faster experiments.
It has been verified by experiments that reducing the training data will lead to a slight decrease in the IoU of moving objects, but the effectiveness of each module is still guaranteed.

\textbf{Evaluation Metrics.}
Following the protocols of LMNet~\cite{chen2021ral}, for quantifying the MOS performance, we measure the Jaccard Index or intersection-over-union (IoU) metric~\cite{everingham2010ijcv} over moving objects, which is given by
\begin{align}
	\text{IoU} = {\text{TP}}/({\text{TP} + \text{FP} + \text{FN}}), \label{eq:miou}
\end{align}
where $\text{TP}$, $\text{FP}$, and $\text{FN}$ represent true positive, false positive, and false negative predictions for the moving class.
 
\textbf{Baselines.} Because there are not many existing learning-based implementations for LiDAR-based MOS available, we only choose three typical approaches using different types of inputs as representatives.
(1)~Range Image View: LMNet~\cite{chen2021ral} uses the residual images as additional channels together with the range images as input to a range image backbone and is trained with binary labels. A kNN post-processing~\cite{milioto2019iros} is used to reduce artifacts on objects' borders. And we choose the best setting of LMNet (with SalsaNext backbone) for comparison.
(2)~Bird’s Eye View (BEV): LiMoSeg~\cite{mohapatra2021limoseg} uses two successive LiDAR scans in 2D BEV representation to perform pixel-wise classification and can run at high frame rates on embedded platforms.
(3)~Point-Voxel View: Cylinder3D~\cite{zhu2021cylinder3D} uses cylindrical partition and point-level feature extractor to segmentation. We modify its open source code\footnote{\url{https://github.com/xinge008/Cylinder3D}} to input two consecutive aligned frames and train it from scratch with MOS-labels to perform MOS.

Since the implementation of LiMoSeg is not publicly available, we report the results from the original paper. The results of Cylinder3D and LMNet are from the retrained models using the same setup as used by our methods.

\textbf{Protocols.} We follow the protocols of LMNet~\cite{chen2021ral}, using the official dataset split to train and validate the network, using 8 residual images as the input of Enc-M and the range image with 5 input channels $(x, y, z, r, e)$ as the input of Enc-A as described in~\secref{sec:range}. 
The generation of the residual image is in line with~\cite{chen2021ral}.
Note that, our method \emph{does not} use any semantic information of different classes to refine predictions, such as vehicle and buildings. This means that our method only needs the binary moving/non-moving labels for training.

\subsection{Evaluation Results and Comparisons} \label{sec::sota_comparison}

\begin{table}[!t]
    \caption{Evaluation and comparison of moving objects IoU on the validation set (seq08) and the benchmark test set.}
    \label{tab:comapre_baseline}
    \centering
    \setlength{\tabcolsep}{8pt}
    \begin{tabular}{lcc|C{1.3cm}|C{1.3cm}}
    \toprule
    \textbf{Methods} & kNN & road & \textbf{validation} & \textbf{test}\\
    \midrule
    LiMoSeg{$^{*}$}\cite{mohapatra2021limoseg} & & & 52.60 & -\\
    Cylinder3D{$^{*}$}~\cite{zhu2021cylinder3D} & &  & 66.29 & 61.22\\

    LMNet~\cite{chen2021ral} & &  & 58.11 & 50.18\\
    LMNet~\cite{chen2021ral} & \cmark &  & 62.51 & 54.54\\
    LMNet{$^{*}$}~\cite{chen2021ral}& \cmark & & 63.82 & 60.45\\
    \midrule
    Ours-v1  & &  & 63.17 & 60.21 \\
    Ours-v1  & \cmark &  & 68.07 & 62.53  \\
    Ours-v1  & \cmark & \cmark  & 66.93  & 69.27  \\
    \midrule
    Ours-v2      & &  & \textbf{71.42} & 64.86 \\
    Ours-v2  &        & \cmark & {69.28} & \textbf{70.16} \\
    \bottomrule
    \multicolumn{5}{l}{\scriptsize{$^{*}$ indicates all frames in training split are used, without downsampling.}}
    \end{tabular}%
    \vspace{-0.2cm}
\end{table}

The 3D LiDAR-MOS evaluation results of moving objects IoU are shown in \tabref{tab:comapre_baseline}.
All the methods are evaluated on both, the validation set (sequences 08), which is unseen during training, and the hidden test split (sequences 11-21) of the benchmark dataset.
The implementation of the bird-eye-view method LiMoSeg is not publicly available, and only the validation set result is reported in the original paper~\cite{mohapatra2021limoseg}.
Hence, its test result is missing.
For LMNet, we use its released code also with our sampled data protocol to retrain the model from scratch, which performs slightly worse than that reported in the original paper.
For a deeper insight, we report the results of our method using only the 2D segmentation network structure called ``Ours-v1", and the complete structure with the proposed PointHead called ``Ours-v2".

As can be seen in~\tabref{tab:comapre_baseline}, our method achieves significantly better performance than the state-of-the-art learning-based methods in terms of LiDAR-MOS.
The improvements come from our designed dual-branch structure with the motion-guided attention and the coarse-to-fine scheme with the PointHead module.
Our method using only the range-image backbone without kNN (Ours-v1) already outperforms most of the baseline methods and is on par with Cylinder3D, which is a dense point cloud-based semantic segmentation method but cannot achieve real-time performance due to the large amount of computation.
When training with the proposed extra KITTI-road sequences, our method gains a better generalization and LiDAR-MOS performance on the hidden test set.
Our method with the proposed PointHead (Ours-v2) achieves the state-of-the-art LiDAR-MOS performance with an IoU of~$70.16\%$ in the SemanticKITTI-MOS benchmark, by exploiting and fusing both the range image and point cloud representations of the LiDAR data. 
Moreover, our method could also run online, which will be further discussed in~\secref{sec:runtime}.

\begin{figure*}[!ht]
    \centering
    \includegraphics[width=0.99\linewidth]{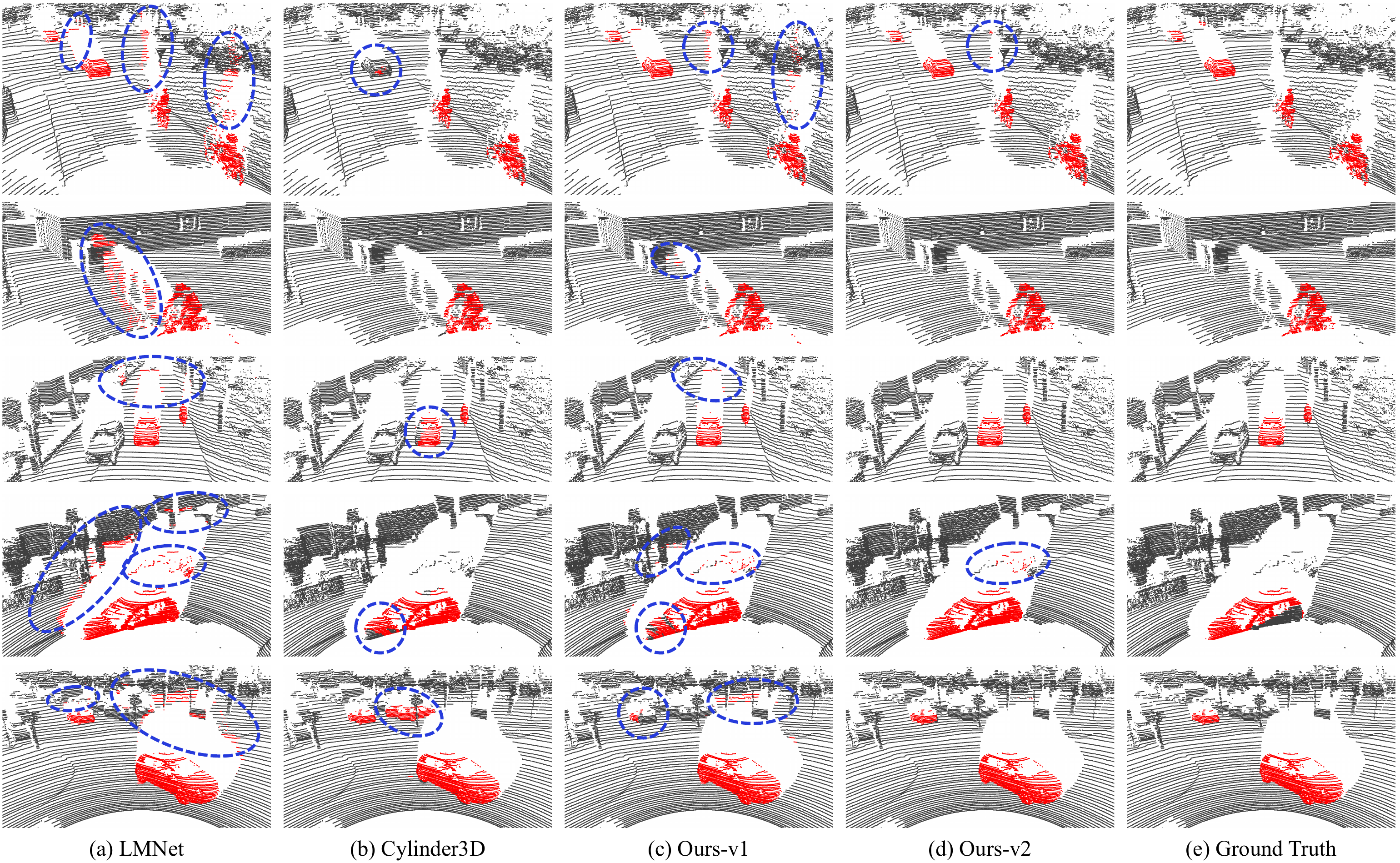}
    \caption{Qualitative results of different methods for LiDAR-MOS on the validation set of the SemanticKITTI-MOS dataset. Blue circles highlight incorrect predictions and blurred boundaries. Best viewed in color and zoom in for details.}
    \label{fig:qualitative_vis}
\end{figure*}

More qualitative comparisons are shown in~\figref{fig:qualitative_vis}. As illustrated, the range image-based LMNet generates many wrong predictions on the borders of the objects, while the point cloud-based method Cylinder3D often misses detecting parts of the moving objects. Using the devised dual-branch structure and motion-guided attention and fusing both different representations of LiDAR scans, our method detects most points belonging to moving objects without bringing artifacts on the objects' borders.

\subsection{Ablation Study} \label{sec:ablation}

\begin{table*}[t]
    \caption{Ablation study of components on the validation set (seq08). ``$\Delta$" shows the improvement compared to the vanilla baseline~(\textit{a}).} \label{table:ablation_study}
    \centering
    \small
        \begin{tabular}{c|l|cc|cc|cc}
        \toprule
         & Baseline and components & w/o kNN & $\Delta$ & w/ kNN & $\Delta$  & w/ PointHead & $\Delta$\\ 
        \midrule
        \rowcolor{lightgray} (\textit{a}) & LMNet (with SalsaNext)        & 58.11 &    & 62.51 &   & 64.05 &  \\
        (\textit{b}) & + DualBranchWithMGAtten  & 60.85 & +2.74   & 65.50 & +2.99 & 67.74  & +3.69\\
        (\textit{c}) & + DualBranchWithMGAtten + SoftPool & 61.65 & +3.54   & 66.27 & +3.76 & 68.52 & +4.47\\
        (\textit{d}) & + DualBranchWithMGAtten + MetaKernel & 62.75 & +4.64  & 67.57 & +5.06 & 70.26 & +6.21\\
        (\textit{e}) & + DualBranchWithMGAtten + MetaKernel + SoftPool & 63.17 & +5.06  & 68.07 & +5.56 & 71.42 & +7.37\\
        \toprule
        \end{tabular}
    \vspace{-0.2\baselineskip}
\end{table*}

\begin{table}[t]
    \caption{Ablation on extra data and different validation setups.}
    \label{tab:ablation_kitti_road}
    \centering
    \begin{tabular}{c|lcc|cc|c}
    \toprule
    & & & & \multicolumn{2}{c|}{\textbf{validation}} &\\
    & \textbf{Methods} & kNN & road & seq08 & seq08+road & \textbf{test}\\
    \midrule
    (\textit{i})& LMNet~\cite{chen2021ral} & \cmark  &  & 62.51 & - & 54.54\\
    (\textit{ii})& LMNet$^{*}$~\cite{chen2021ral} & \cmark  &  & 63.82 & - & 60.45\\
    (\textit{iii})& Ours-v1 & \cmark &  & 68.07 & - & 62.53\\
    \midrule
    (\textit{iv})& LMNet~\cite{chen2021ral}  & \cmark & \cmark  & 54.26 & 81.80 & 62.98\\
    (\textit{v})& Ours-v1 & \cmark & \cmark & 66.93 & 84.67 & 69.27\\
    \bottomrule
    \multicolumn{7}{l}{\scriptsize{$^{*}$ indicates all frames in training split are used, without downsampling.}} 
    \end{tabular}%
    \vspace{-0.5\baselineskip} 
\end{table}

In this section, we conduct several ablation experiments on the validation set (sequence 08) of the SemanticKITTI-MOS dataset to analyze the effectiveness of different components of our method.
For the validation of each setup, we train 3 times and report the averaged results.

We first provide an ablation study on the architecture of the proposed network in~\tabref{table:ablation_study}.
We vertically compare each module with different setups of our proposed network to the vanilla LMNet (with SalsaNext backbone)~(\textit{a}).
``$\Delta$" refers to the improvement gained by different setups compared to the baseline~(\textit{a}).
We report the results of using the proposed dual-branch architecture with motion-guided attention (MGAtten.) under (\textit{b}). 
The dual-branch structure with attention improves the vanilla method (w/o kNN) by $2.74\%$ in terms of IoU.
On this basis, we add the Meta-Kernel convolution (\textit{c}), SoftPool (\textit{d}), and combine them together (\textit{e}). The performances further improve consistently.
Our final setup gains an improvement of $5.06\%$ compared to the baseline without kNN.

We also divide different setups horizontally into three groups: without using a kNN post-processing (w/o kNN), with a kNN post-processing (w/ kNN), and with using our proposed PointHead (w/ PointHead).
When comparing the setups with or without using kNN for post-processing, we see a clear improvement after applying just a simple kNN in all setups.
Instead of using a kNN, we propose to use a PointHead to further refine the MOS results of the 2D range image network by fusing the supervision from 3D points level. As can be seen, using our proposed PointHead, we obtain even better results than both with and without kNN in all setups. Different to kNN as an extra post-processing module, our PointHead is a part of the network and enables our method to exploit and fuse both representations of LiDAR scans in a more elegant end-to-end manner. 
More specifically, compared to the same setting without using kNN post-processing, our proposed PointHead achieves a maximum absolute $3.35\%$ improvement over its counterpart using a kNN and $8.25\%$ improvement over its counterpart without using post-processing.
As shown in~\figref{fig:motivation} and~\figref{fig:qualitative_vis}, the qualitative results also prove that PointHead can better handle the blurred boundary of moving objects.
Filtering based on kNN votes is limited by the receptive field (size of k). 
In contrast, our module is end-to-end trained, and the receptive field is more flexible, so it performs better than with kNN post-processing.

Another ablation study about using extra training data from the KITTI-road sequences is shown in~\tabref{tab:ablation_kitti_road}. In the original SemanticKITTI dataset, sequence 08 is set as the validation set considering that it contains all the different categories of semantic classes. 
However, for MOS task, we found that sequence 08 is difficult to represent all different situations to well evaluate the trained models for MOS, since sequence 08 only contains the LiDAR data collected in the urban environment. Therefore, we introduce extra KITTI-road data to provide more training and validation data collected in different environments such as country roads or highways. 

As shown in~\tabref{tab:ablation_kitti_road}, we test both LMNet and our methods using different training setups. As can be seen, the models trained using extra KITTI-road data, (\textit{iv}) and (\textit{v}), achieve better performance on the hidden test set compared to their counterparts, (\textit{ii}) and (\textit{iii}) trained only using SemanticKITTI-MOS data.
This indicates that our trained and validated models achieve good generalization ability using the proposed additional data.
Interestingly, the performance of the models trained with extra KITTI-road data decreased on the original validation set sequence 08, which indicates that the performance/improvement of the validation and test sets are not strictly positively correlated.

\subsection{Runtime and Efficiency} \label{sec:runtime}
The runtime is evaluated on sequence 08 (about 122k points per scan) with Intel Xeon Silver 4210R CPU @ 2.40\,GHz and a single NVIDIA RTX 3090 GPU.
In Table~\ref{tab:runtime}, we report the average time comparison of several baseline methods and ours.
Due to serial processing, the PointHead block affects the total time. 
We also provide a lightweight PointHead-Lite ($^{\dag}$) with fewer layers to trade-off the accuracy and speed, which is faster than the frame rate 10\,Hz of a typical 3D LiDAR sensor (\eg, Velodyne or Ouster).

\begin{table}[!t]
    \caption{Comparison of running time (ms) with baseline methods.}
    \centering
    \label{tab:runtime}
    \begin{tabular}{c|c|c|c|c}
    \toprule
    LMNet & Cylinder3D & Ours-v1 & Ours-v2 & Ours-v2{$^{\dag}$} \\
    \midrule
    13.31 & 124.38 & 41.81 & 116.93 & 85.25 \\
    \bottomrule
    \end{tabular}
    \vspace{-0.2cm}
\end{table}

\section{Conclusion}
\label{sec:conclusion}
In this paper, we have presented a novel and effective network for LiDAR-based online moving object segmentation. 
Our method uses a dual-branch structure to better explore and fuse the spatial and temporal information that can be obtained from sequential LiDAR data. A point refinement module is designed to significantly reduce the boundary-blurring artifacts of the objects, and this coarse-to-fine strategy enables our method to operate online.
We additionally annotated the KITTI-road dataset to enrich the training data, which enhanced the generalization ability of the model. 
Experimental results on the SemanticKITTI-MOS dataset demonstrate the state-of-the-art performance of our proposed method.

\bibliographystyle{ieeetr}

\footnotesize
\bibliography{glorified,new}

\end{document}